%% file: root.tex
\documentclass[conference]{IEEEtran}
\IEEEoverridecommandlockouts
\usepackage{cite}
\usepackage{amsmath,amssymb,amsfonts}
\usepackage{algorithmic}
\usepackage{graphicx}
\usepackage{textcomp}
\usepackage[hidelinks]{hyperref}
\usepackage{xcolor}
\usepackage{tabularx} 
\usepackage{algorithm}
\graphicspath{{./figs/}}

\def\BibTeX{{\rm B\kern-.05em{\sc i\kern-.025em b}\kern-.08em
    T\kern-.1667em\lower.7ex\hbox{E}\kern-.125emX}}
\begin{document}

\linespread{0.85}

\title{MTAC: Hierarchical Reinforcement Learning-based Multi-gait Terrain-adaptive  Quadruped Controller}
\author{Nishaant Shah, Kshitij Tiwari, and Aniket Bera%
\thanks{The authors are with the Department of Computer Science, Purdue University, USA,
        {\tt\small \{shah971,tiwarik,ab\}@purdue.edu}}
        }

\maketitle

%% Main paper sections
\input{sections/00abstract}
\input{sections/01intro}

\input{sections/02related}
\input{sections/03method}

\input{sections/04evaluation}
\input{sections/05conc}

%% Bibliography
\bibliographystyle{IEEEtran}
\bibliography{root.bib}

\end{document}

%% file: sections/00abstract.tex
\begin{abstract}
Urban search and rescue missions require rapid first response to minimize loss of life and damage. Often, such efforts are assisted by humanitarian robots which need to handle dynamic operational conditions such as uneven and rough terrains, especially during mass casualty incidents like an earthquake. Quadruped robots, owing to their versatile design, have the potential to assist in such scenarios. However, control of quadruped robots in dynamic and rough terrain environments is a challenging problem due to the many degrees of freedom of these robots. Current locomotion controllers for quadrupeds are limited in their ability to produce multiple adaptive gaits, solve tasks in a time and resource-efficient manner, and require tedious training and manual tuning procedures. To address these challenges, we propose MTAC: a multi-gait terrain-adaptive controller, which utilizes a Hierarchical reinforcement learning (HRL) approach while being time and memory-efficient. We show that our proposed method scales well to a diverse range of environments with similar compute times as state-of-the-art methods. Our method showed greater than 75\% on most tasks, outperforming previous work on the majority of test cases. 
\end{abstract}

%% file: sections/01intro.tex
\section{Introduction}
The past few decades have witnessed a high number of natural and man-made disasters. In the wake of these natural disasters and mass casualty incidents such as earthquakes or building collapses, hazardous conditions make these environments extremely difficult to access for first responders. Moreover, the potential presence of asbestos, dust, extreme temperatures, and unstable structures endanger the lives of first responders. In the search for advanced technologies to alleviate these problems, mobile robots have been increasingly discussed as potential solutions. Robots have been used to assist first responders in recent devastation such as the collapse of the World Trade Center, the 2004 Mid-Niigata earthquake in Japan, the 2005 Hurricanes Katrina, Rita, and Wilma in the United States, as well as the 2011 Tohoku earthquake and tsunami in Japan \cite{liu2012Learning}. Nevertheless, the challenge of robot-assisted urban search and rescue is far from being solved. In order to fully explore post-MCI environments, robots must be able to navigate dynamic obstacles, as well as cluttered and unstructured terrains. 
In recent years, tracked robots have been used to drive across uneven terrain. For, in response to the Great Eastern Japan Earthquake, tracked robots such as the KOHGA3 were used for inspection of disaster sites \cite{matsuno2014Utilization}. Although this teleoperated robot was able to drive over uneven ground and use its sensors to provide the operators with information about the site, it was limited in its ability to climb stairs or steep slopes. 
%% Figure for motivation/visual abstract
\begin{figure}[!htbp]
    \centering
    \includegraphics[scale=0.2]{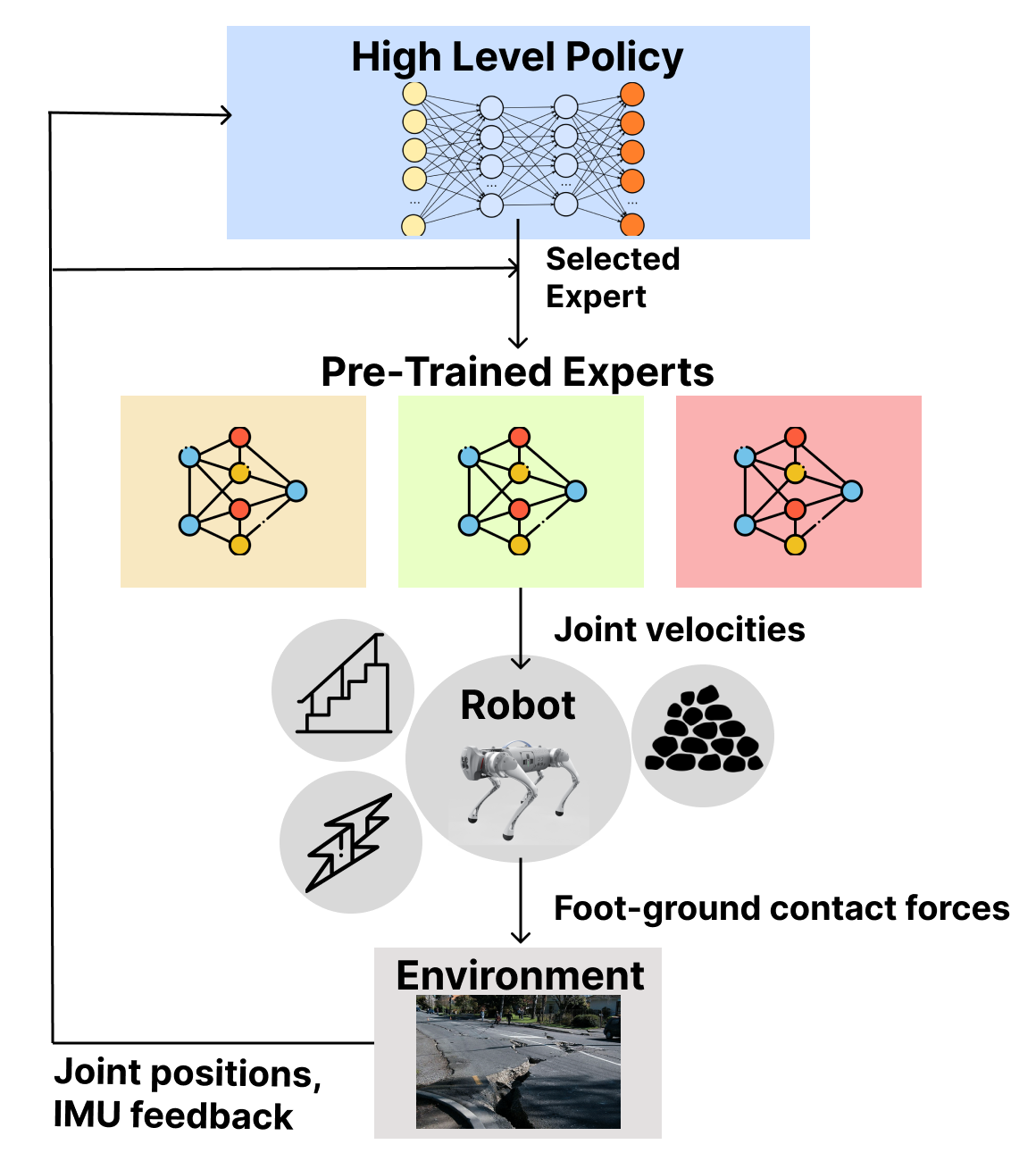}
   \caption{\textit{The proposed controller solves rough terrain navigation using hierarchical learning. A high-level policy is trained to act over a family of pre-trained low-level experts trained to execute unique gaits. Together, these policies are able to execute adaptive locomotion on unstructured terrains}}
    \label{fig:visual abstract}
    \vspace{-20pt}
\end{figure}
Quadruped robots, thanks to the high degree of versatility that comes with their legged design, have great potential to navigate complex environments that their counterparts, such as wheeled or tracked robots, cannot \cite{bledt2018cheetah,hutter2016anymal,katz2019mini}. Because quadrupeds use legs to move around, they have the freedom to choose the points at which they make contact with their surroundings and quickly change their footholds based on their environment. These features give quadruped robots the potential to perform agile locomotion that previously only animals were capable of \cite{hwangbo2019Learning}. 
Despite their highly versatile design, however, the control of these legged systems has proven to be a challenging task. Although their physical design is bio-inspired, existing control algorithms for quadrupeds struggle to replicate animals' natural ability for dynamic locomotion over diverse terrains. Locomotion control for quadrupeds in rough terrain is a challenging problem, because it requires the robot to adapt its gait to diverse conditions, and the resources on these mobile robots are constrained \cite{tiwari2018estimating,Tiwari2019Unified}. Currently, both conventional model-based controllers, as well as learning-based model-free controllers, have shown some success in the field of quadrupedal locomotion for rough terrain. Nevertheless, they have limitations when it comes to addressing a diverse set of environmental conditions and resource efficiency. Furthermore, existing controllers typically converge to a standard trotting gait, which is not always optimal for certain velocity ranges or all terrains \cite{hwangbo2019Learning,tan2021Hierarchical}. In fact, specific gaits are more optimal for certain velocity ranges than others \cite{kim2021Learning}. Multiple gait modes allow the robot to switch between gaits, thereby performing more complex and dynamic motions, such as changing from walking to pronking \cite{surana2023Efficient,bellicoso2018Dynamic}. It has been shown that behaviors learned from scratch are more energy efficient for locomotion tasks and capable of more complex locomotion compared to primitive gaits like trot \cite{rudin2022Advanced}. Together, these works demonstrate the need for a learning-based controller that can illicit multiple emergent gaits from the robot, while accomplishing tasks in a time and energy-efficient manner. 
In this paper, we propose the use of hierarchical reinforcement learning (HRL) to train a multi-gait terrain-adaptive locomotion controller for quadruped robots (see Fig.~\ref{fig:visual abstract}). To this end, we make the following novel contributions:
\begin{itemize}
    \item  The policies are trained end-to-end, without using human-engineered trajectory generators (TG) or motion primitives. We \textbf{train our locomotion controller from scratch}, using the design of the terrain in the simulated training curriculum to elicit different gaits rather than developing gaits from motion primitives. The resulting gaits are more natural and diverse because they do not draw inspiration from existing motions. These gaits are referred to as \textit{emergent behaviors}, and they are specialized for the conditions in which they are trained. This yields more efficient performance than motion primitives that are generalized to multiple terrain types. In this work, we focus on training environments that contain stairs, gaps, and uneven ground. These three distinct environments elicit unique behaviors from their respective policies. Training the high-level policy to modulate over these pre-trained policies allows for these expert skills to be synthesized within a single navigation task/mission.
    \item The \textbf{model reduces navigation time} by generating multiple emergent gaits and performing at high velocities. The methodology allows for a reduction of navigation time, by training a family of complimentary and unique gaits. Previous works have shown that certain gaits are more efficient and appropriate for certain velocity ranges, and terrain types. By straying from traditional pre-determined gaits and eliciting emergent behaviors from our low-level expert policies, our proposed controller reduces the time taken to complete a navigation task/mission. In doing so, we extend the potential range of the robot for a single mission, without the need for external batteries, which would be a cumbersome addition to the quadruped's payload. This is necessary for applications in urban search and rescue, such as earthquakes, when missions must be completed quickly- especially due to the high chance of after-shocks which could further compromise the integrity of buildings and human lives. 
    % \item The controller runs on a \textbf{lightweight sensor suite}, allowing for deployment on resource-constrained robots. The model takes only proprioceptive inputs from the robot's joint encoders as inputs. This means that the method can be deployed on robots without any additional sensors or hardware being installed. By making the navigation sensor suite minimalist, we allow more room for other sensors to be mounted. During USAR scenarios, quadrupeds will need additional hardware mounted in order to complete tasks such as damage assessment, algorithmic triage, or victim identification. For this reason, we make our navigation platform resource-constrained so that there is more room to integrate other capabilities on top of our navigation platform. 
\end{itemize}

%% file: sections/02related.tex
\begin{figure}[!htbp]
    \centering
    \includegraphics[scale=0.57]{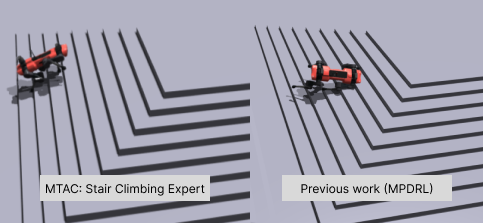}
    
    \caption{\textit{Shown above is MTAC's pre-trained stair expert successfully navigating a difficult grade stair pyramid, as a model trained through the previous work's generalized curriculum falls on a lower difficulty level.}}
    \label{fig:comparison}
        \vspace{-15pt}
\end{figure}
\section{Related Works}

This section discusses the varied control methods adopted for quadrupedal locomotion, focusing on model-based controls, impact-aware locomotion, and learning-based methodologies.

\subsection{Quadrupedal Locomotion}\label{subsec:quad-locomotion}

Quadrupeds, with their myriad degrees of freedom, pose significant control challenges, especially when aiming for agile motion. The literature reveals a range of strategies that address this intricacy.

\subsubsection{Model-based Control}

Traditional approaches in this category leverage motion planning algorithms, deeply rooted in the robot's dynamics \cite{kalakrishnan2010Fast,bellicoso2018Dynamic,lee2019Whole}. These approaches usually employ a blend of state machines and meticulously crafted foot trajectories. They consider various parameters, including potential slippage and ensuring appropriate ground contact \cite{jenelten2019Dynamic}. While these methods demonstrate effectiveness, they sometimes falter in dynamic scenarios where real-time adaptability is paramount. Notably, Fankhauser et al. introduced a more adaptive controller. It continuously maps the robot’s environment, adjusting to changes such as new obstacles. However, it does have limitations in recognizing deformable surfaces or accounting for terrain alterations as the robot moves \cite{fankhauser2018Robust}.

\subsubsection{Impact-aware control}

The evolution of robotic hardware has sparked interest in controllers that integrate direct feedback from the robot's appendages. Guo et al.'s work, which introduced a sensorized foot pad capable of terrain classification, exemplifies this shift \cite{guo2020Soft}. By merging such sensor feedback with existing control algorithms, there's a marked improvement in how robots adapt to terrains \cite{valsecchi2020Quadrupedal,bellicoso2017Dynamic}. A noteworthy development in this domain is Zhang et al.'s risk assessment network, which employs a combination of elevation maps and learning policies to navigate rough terrains. While promising, this model acknowledges some limitations in teaching agile behaviors \cite{zhang2021TerrainAware}.

\subsubsection{Learning-based control}

The increasing dominance of reinforcement learning in robotic control is evident in quadrupedal locomotion as well. Policies Modulating Trajectory Generators (PMTG) is a pioneering architecture in this space, where neural networks determine trajectory parameters based on robot states \cite{pmlr-v87-iscen18a}. The Evolved Environmental Trajectory Generators (EETG) framework builds upon PMTG. It introduces a level of diversity in trajectory generation, even if its primary results are limited to simulations \cite{surana2023Efficient,DBLP:journals/corr/MouretC15}. Real-world implementations of PMTG-like architectures have seen success, especially Tan et al.'s work that exhibited robustness across varied terrains \cite{tan2021Hierarchical}. In contrast, Lee et al. emphasized training under diverse simulated conditions, though their results showed a preference for a single gait \cite{doi:10.1126/scirobotics.abc5986}. Finally, Rudin et al. took a different track, advocating for a holistic end-to-end approach. This method demonstrated unique gaits and better energy efficiency, but it had its set of challenges, especially in reverse locomotion \cite{rudin2022Advanced,schulman2017proximal}.

%% file: sections/03method.tex
\section{Multi-gait Terrain-adaptive Controller (MTAC)}
In this section, we first present an overview of the proposed multi-gait controller MTAC (Sec.~\ref{subsec:overview}) followed by detailed descriptions of the corresponding high and low-level controllers in Sec.~\ref{subsec:high-level-controller} and Sec.~\ref{subsec:low-level-controller}, respectively.

\begin{figure}[!htbp]
    \centering
    \includegraphics[scale=0.25]{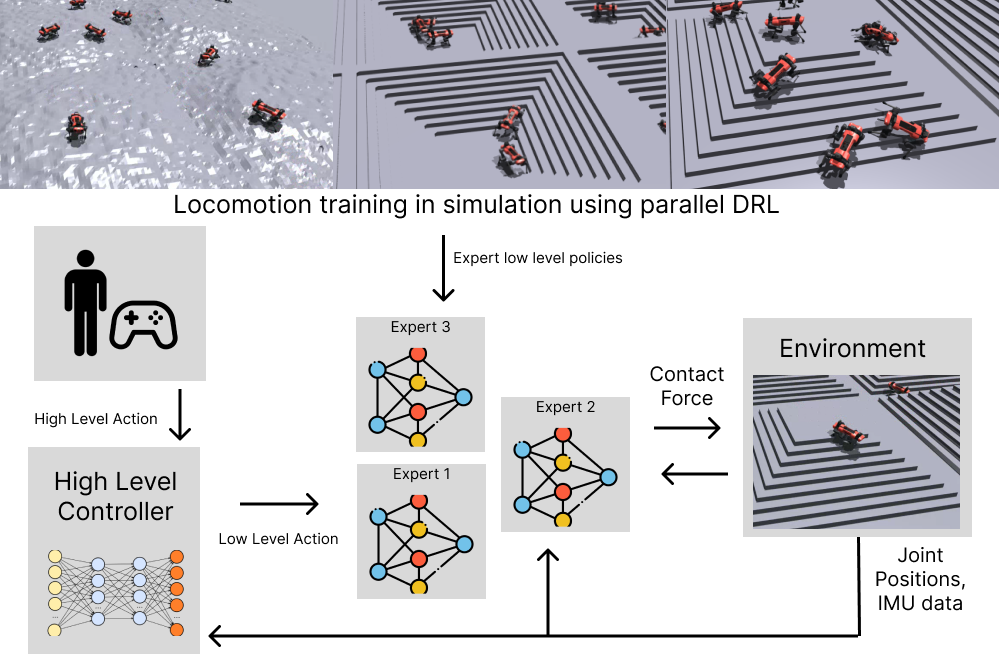}
    \caption{\textit{The methodology for MTAC is pictured above. At the highest level, is the input of a high-level action into the controller. Based on this command, the high-level policy then selects a low-level action to be executed by one of the 3 expert policies. The training process for these experts is also pictured. Then the robot executes the low-level action and interacts with the environment. These contacts then start the next iteration of the feedback loop.}}
    \label{fig:method}
\end{figure}

\begin{algorithm}\label{alg:alg1}
   \algorithmicwhile{} controller running \algorithmicdo{}
       \\Get command velocity 
       \\Get observation \(o^{high}_t\)
       \\\(a^{high}_t\), \(d\) = \(\pi^{high}(a_t|o_t)\) 
      \\\algorithmicwhile{} \(t_{elapsed} < d\)
            \\Get observation \(o^{low}_t\)
            \\Run \(\pi^{low}\)
\caption{Locomotion Control Algorithm} 
\end{algorithm}

\subsection{Overview}\label{subsec:overview}
The proposed controller splits the task of locomotion into multiple sub-tasks using a hierarchical learning architecture. An overview of the method is shown in Fig. \ref{fig:method}. The Multi-gait Terrain-Adaptive Controller (MTAC)\footnote{\href{https://ideas.cs.purdue.edu/research/QuadrupedLocomotion}{https://ideas.cs.purdue.edu/research/QuadrupedLocomotion}} consists of a high-level control policy, which regulates a family of low-level policies trained to exhibit diverse emergent skills. These skills are optimized for a range of sub-tasks. Each low-level policy becomes an expert in certain skills that are optimized for its respective sub-task. The high-level policy selects which expert to apply to unseen environments with the goal of successfully navigating to a desired goal. This method is tested in cluttered and unstructured environments in simulation. The execution of the controller is described in Algorithm~\ref{alg:alg1}. First, the controller receives a commanded velocity vector for the robot's base. Then an observation of the robot's current state is made. Based on this observation,  the high-level policy outputs a high-level action and the estimated duration of this action. The high-level action activates a low-level gait mode, which controls the motion of the legs of the robot as it traverses the terrain.
\begin{algorithm}\label{alg:alg2}
Load environment \(env\)\\
Send command \(c\) 
 \algorithmicor ~\(t_{elasped}\) \(<\) 25 \algorithmicdo{}
    \\Get observations \(o^{high}_t\), \(o^{low}_t\)
    \\ \(a^{high}_t\)= \(\pi^{high}(a_t|o_t)\)
    \\\(\pi_{expert}\) =  \(a^{high}_t\)
    \\ \(a^{low}_t\) =  \(\pi_{expert}(a_t|o_t)\)
    \\\(env\).increment-step(\(a^{low}_t\))
    \\Calculate reward \(R(t)\)
    \\Pass reward to \(\pi_{high}\)

    \caption{High Level Controller Training}
\end{algorithm}
\begin{algorithm}\label{alg:alg3}
Load environment \(env\)\\
\algorithmicwhile{} \(t_{elapsed}\) \(<\) 20 
 \algorithmicor ~\(n_{steps}\) \(<\) 25 \algorithmicdo{}
    \\Get observation \(o^{low}_t\) 
    \\ \(a^{low}_t\)= \(\pi^{low}(a_t|o_t)\)
    \\\(env\).increment-step(\(a^{low}_t\))
    \\Calculate reward \(R(t)\)
    \\Pass reward to \(\pi_{low}\)

    \caption{Low Level Controller Training}
\end{algorithm}

 \subsection{High Level Controller}\label{subsec:high-level-controller}
One of the novel components of our approach, MTAC, is the high-level controller which acts as the \textit{policy over policies}. Instead of a single policy controlling locomotion, the high-level controller is trained using DRL to allocate certain periods of locomotion to the more specialized expert policies. We represent the problem of locomotion control in discrete time steps. At each time step $t$, the robot obtains an observation \(o_t \in O\), performs an action \(a_t \in A\), and receives a reward \(r_t \in R \). The robot chooses  \(a_t\) based on a policy \(\pi(a_t|O_t)\). We use deep reinforcement learning (DRL) to solve this control problem. Using DRL, the goal is to obtain a policy \(\pi^*\) that maximizes the sum of the discounted rewards over an infinite time horizon. In order to implement the DRL framework, we first formulate the high-level task of locomotion as a Markov decision process (MDP). MDP is a mathematical tool used in reinforcement learning for modeling discrete-time decision-making problems that are partially controllable and partially stochastic. An MDP is defined by the tuple \((A, S, p, r)\), where $A \in \mathbb{R}^{3}$ represents the continuous action space, $S$ where $S \in \mathbb{R}^{15}$ represents the continuous state space, $p$ is the unknown state-transition probability, and the reward is $r$. This tuple is stored in memory in a replay buffer so that the policy is able to keep track of its history. For the high-level controller, we define a continuous action space consisting of a confidence rating for the performance of each pre-trained expert policy. The expert with the highest confidence is then selected to control the robot's gait. For this controller, the state space or observation consists of the quadruped's 12 joint positions and velocities, as well as the base twist, orientation, and direction command. The training procedure for the high-level controller is presented in Algorithm \ref{alg:alg2}. 

 \subsection{Low Level Control}\label{subsec:low-level-controller}
For the low-level controllers, we train a family of expert policies. Each expert policy consists of a neural network trained using the proximal policy optimization (PPO) algorithm \cite{schulman2017proximal}. By using PPO, we take advantage of entropy regularization to encourage the robot to explore a wider range of behaviors or take actions with more entropy. As done with the high-level controller, we also formulate the problem of low-level control as an MDP. We define the continuous action space$A \in \mathbb{R}^{D}$ as a set of joint reference angles for each of the robot's actuated joints, D being the number of degrees of freedom of the particular quadruped. These desired joint states are then fed into the robot's joint PD controller. The controller's observation space is defined by the robot's joint positions and velocities as well as base twist, orientation, and direction command. For training the experts, we expand on the massively parallel training procedure presented in \cite{rudin2021Learning} as this work is open source and allows for rapid training of legged locomotion policies, by simulating thousands of robots in a single environment and learning a policy in a parallel manner. During training, domain randomization is applied in the form of friction randomization, mass randomization, and randomized disturbances to the robot's base. 
 \begin{figure}[!htbp]
    \centering
    \includegraphics[width=\columnwidth,height=5cm]{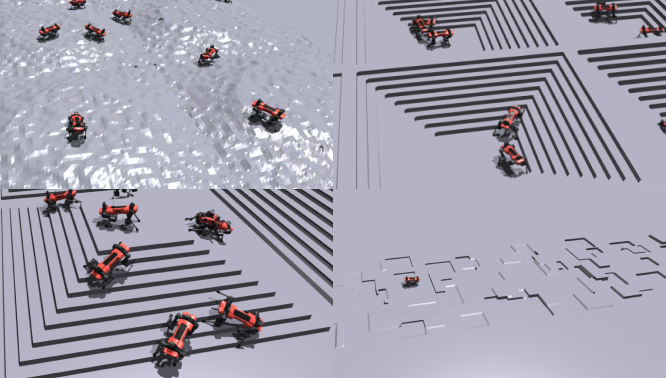}
    \caption{\textit{Expert policies are developed by training in specialized environments that consist of focused terrain types. Here we show three environments: bumpy terrain (top left), stair pits (top right), stair pyramids (bottom left) and steps (bottom right). All of these terrain types are parameterized so that difficulty can be gradually varied across the map.}}
    \label{fig:curriculum}
        \vspace{-12pt}
\end{figure}

 We train three expert policies using a specialized terrain curriculum with increasing difficulty at each level within the map. By doing so, we allow each policy to learn a subset of skills that is more specialized towards its respective environment. Each simulation environment is shown in Fig.~\ref{fig:curriculum}, and described in detail later in this paper. Each policy is trained in its respective environment with the same process. First, A 10 by 10 grid is generated consisting of the environment's respective terrain. The grid is arranged so that the terrains of varying difficulty are presented. In each row, the parameters of the terrain are incremented in order to make the terrain more difficult to traverse. Then the robots are placed in the simulation in large clusters in the middle of each grid square. Short episodes of 20 seconds are implemented, and an always positive rewards scale is implemented in order to avoid early termination. At the end of each episode, agents are moved to other points on the grid based on performance, so that over the course of the training period, all of the agents will first learn from the easy examples of the terrain, before learning to address more severe and difficult terrains. One training iteration contains 25 steps, and 1500 iterations of this process are performed. After 1500 iterations, a final policy is generated. In training, the maximum episode length is 20 seconds, and episodes last over multiple iterations and, therefore, multiple policy updates. The low-level training procedure is detailed in \ Algorithm~\ref{alg:alg3}. As opposed to the previous work, which aims to improve performance by approaching the upper limit of robots simulated, we reduce the number of robots per environment from 4096 to 1024, allowing for training to be more lightweight, while still converging on more specialized behaviors. Furthermore, by training each expert in its own environment, we increase the ceiling for the maximum difficulty of terrain faced by the agents in training. We also focus more robots on a single locomotion task at a time, rather than splitting their focus across learning multiple types of terrains. 

 \subsubsection{Bumpy Terrain}
In this environment (shown in Fig.~\ref{fig:curriculum} (top left)), the quadruped is trained to follow a goal velocity while traversing a terrain with many random bumps and elevation changes. In this environment, the policy is learned in which the robot can quickly and accurately follow its commanded velocity while maintaining balance and safe footholds. The minimum and maximum height of the irregularities parameterize this terrain. As the difficulty scalar increases, the range of the min and max heights does as well. Domain randomization is applied, consisting of random force applied to the base as well as variations in the mass of the payload of the robot. Here, the quadruped learns a unique gait similar to the trot; however, it includes dynamic movements in the shoulder joints, which allow for more versatile motion in multiple directions. The robot also learns how to walk both forward and backward, as well as a sideways shuffling motion. Unlike the policy trained in stairs, this policy exhibits more economic steps, in which the robot lifts its feet just high enough to step over the irregularities in the ground, without lifting the foot too high and spending unnecessary time and energy.  It can also be observed that this policy executes similar efficient motions when moving laterally as well, by not over-actuating its shoulder joints. The agents also show smooth motion along non-uniform trajectories, such as a smooth transition from lateral motion to forward motion as the direction of the goal velocity changes. 

\subsubsection{Stair pits and pyramids}
In this environment (shown in Fig.~\ref{fig:curriculum} (top right and bottom left, respectively)), the quadruped is trained to follow a goal velocity while climbing up, down, and over pyramids of stairs with increasing difficulty as they move around the map. Both the stair pyramids and stair pits are parameterized in terms of step width, step height, and the size of the platform at the apex. As the robot advances in the training curriculum, it faces pyramids and pits with larger step heights and smaller platform sizes. This gradual introduction of more difficult conditions allows the policy to gradually learn how to locomote over the terrain, allowing for more exploration of the solution space rather than an early introduction of failure. 
In this environment, the robot learns a foot-trapping motion to lift its legs in order to climb over the steps. Additionally, the agents exhibit more independent leg movements on the way down the staircases, as well as sideways movements down the stairs, to maintain better balance. Additionally, the robot replicates the foot-trapping motion in the horizontal axis as well, using its shoulder joint to lift its legs while climbing up the stairs sideways. By being able to move and traverse the terrain in multiple directions, this policy saves navigation time and efficiency, unlike hand-engineered trajectory generators, which do not include backward and sideways motions. These motions are learned with respect to the terrain in which the agent is trained, so unique behaviors such as foot trapping, and horizontal foot trapping are specialized for the task of navigating stairs. When the same model is run in a more general terrain curriculum, these behaviors are lost as the agent tries to converge on a gait that is more applicable to the generalized terrain. Fig.\ref{fig:comparison} shows how these unique behaviors increase performance. 

\subsubsection{Steps}
In this environment (shown in Fig.~\ref{fig:curriculum} (bottom right)), the quadruped is trained to follow a goal velocity while traversing a terrain with randomly placed rectangular steps on the ground. This terrain is parameterized by the height, length, and width of the steps. As the difficulty scalar increases, the dimensions of the steps increase. In this environment, the robot learns to negotiate the terrain in a similar manner to the stair-climbing expert; however, it adapts more to the fluctuating ups and downs. The trained policy in this scenario exhibited a pronking-like behavior, however, with more caution and without the airborne motion typically associated with pronking. Instead, the model learns to synchronize its front and back legs to overcome the height barrier of the presented steps.

%% file: sections/04evaluation.tex
\section{Empirical Evaluation}

\subsection{Quadruped robot platform}
 The ANYmal quadruped is a 12-degree-of-freedom robot, with a torque-controlled actuator on each joint \cite{hutter2016anymal}. When training and evaluating our policies, an accurate physical model of the ANYmal C robot is used in simulation, together with pre-trained actuator networks, to realistically mirror the real-world actuation of ANYmal's joints.

\subsection{Scenario}
We evaluated each of the pre-trained expert policies in environments consisting of their respective terrains and compared the results with those of a policy trained using the same method as \cite{rudin2021Learning}. Each environment consisted of a single row of 5 sections of terrains. For the stair expert, these 5 sections consisted of stairs of varying difficulty. For the bumpy ground expert, the environment consisted of hills of various heights with varying difficulty in bumps. For the step expert, step dimensions of various heights were included. The expert test cases can be seen in Fig. \ref{fig:curriculum}. In each test, the robot is given a forward velocity command, and the completion rate is measured over 10-15 trials. For each test scenario, the robot's performance is evaluated on the terrain of low and high difficulty, as well as with low and high commanded velocity. For low difficulty, the difficulty scalar was reduced by 50\%, this scalar affects the height and severity of the changes in the terrain. For high difficulty, the difficulty scalar was increased to 100\%, while the exact values vary for each task. For high velocity, the robots were commanded to move forward at 1.75 meters per second, and for low velocity, they were commanded to move forward at .75 meters per second. 

\subsection{Results}
\begin{figure}[!htbp]
    \centering
    \includegraphics[scale=0.4]{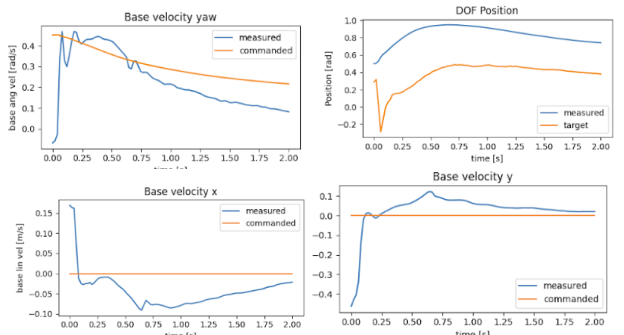}
    \caption{\textit{Metrics from training in the bumpy terrain environment are shown. We can see that the x and y base velocities, as well as the yaw of the robot, converge towards the commanded values as training time progresses. Similarly, as time progresses, the positions of the robot's degrees of freedom also more closely resemble the commanded values.}
    }
    \label{fig:Bumpy-Expert-Plots}
\end{figure}

\begin{figure}[!htbp]
    \centering
    \includegraphics[scale=0.475]{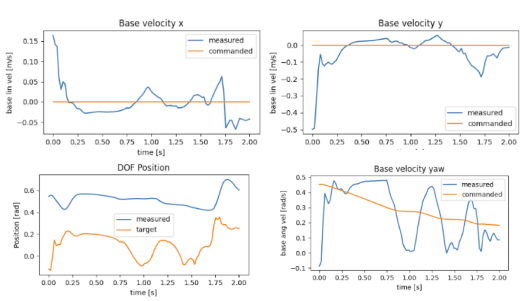}
    \caption{\textit{Metrics from training in the stair pyramid terrain environment are shown. We can see that the x and y base velocities, as well as the yaw of the robot, converge towards the commanded values and oscillate around the endpoint as training time progresses. Similarly, as time progresses, the reference positions of the robot's degrees of freedom also closely resemble the commanded values.}}
    \label{fig:StairPlot}
    
 \vspace{-15pt}
\end{figure}

The results of these experiments are shown for stairs in Table \ref{tab:stair_table}, for bumpy terrain in Table \ref{tab:bumpy_table}, and for stepped terrain in Table \ref{tab:step_table}. In each table the completion rate is represented by the column C.R. After conducting multiple trials, the number of trials in which the quadruped successfully makes it across the terrain is recorded and represented as the Task Completion Rate. Furthermore, the accuracy at which the expert policies are able to track their speed and direction commands as well as the error between the desired joint positions robot's legs and the actual measured values. These results are shown for the bumpy terrain expert in Fig. \ref{fig:Bumpy-Expert-Plots}. Here we can see how this policy converges to closely match the velocity commands as well as closely resembles the targeted joint positions. Similar results are shown for the stair-climbing expert in Fig. \ref{fig:StairPlot}. However, for this policy, the measured velocities show more oscillation and above and below the target curve due to the more drastic changes in slope and elevation for the robot base. The success rate of the policy shows that this is not an issue in terms of performance and rather caused by the nature of the task. As shown in \ref{tab:stair_table}, the MTAC stair expert showed the best results with a high success rate. Similarly, the MTAC step expert also had a high percentage of completion. For the bumpy terrain test, both models performed similarly, performing well on the easier difficulties, and with lower completion rates as the speed and difficulty increased. Nevertheless, in the most difficult test case, with full difficulty for the bumpy terrain and high velocity, MTAC did perform better, with a completion rate more than double that of the previous work. These results show how gaits generated using MTAC's methodology result in greater adaptability to environments with difficulties that were not faced in training, and with navigating difficult terrain at velocity ranges higher than those faced in training. In the case of an MCI, these traits would be more desirable as fast response, and navigation of unpredictable terrain are both necessary in MCI scenarios.

% \begin{figure}[!htbp]
%     \centering
%     \includegraphics[scale=0.3]{figs/TestCase.png}
%     \caption{A one by 5 grid of varying difficulty is used to evaluate the performance of the pre-trained expert low-level controllers. During test scenarios, the robot is commanded with a high forward velocity and the time and rate of completion is measured. Successful completion is considered as reaching the end of the 5 by 1 grid. 
%     }
%     \label{fig:TestCase}
% \end{figure}
\begin{figure}[!htbp]
 \vspace{-15pt}
     \centering
     \includegraphics[scale=0.3]{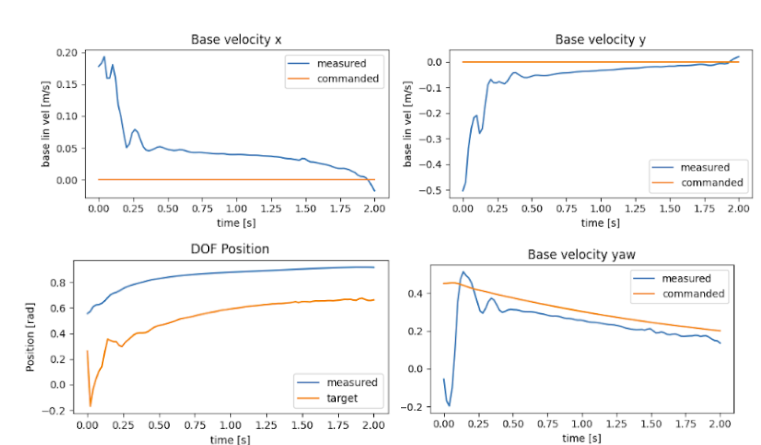}
    \caption{\textit{Here is shown the performance of the MTAC stepping expert during training. It shows how the body velocities converge closely to the commanded velocities. Similarly, the DOF positions also converge to match the desired positions.}
     }
     \label{fig:StepPlot}
         \vspace{-15pt}
 \end{figure}
\begin{table}[]

 \vspace{-15pt}
\caption{Stair pyramid performance metrics. Each model was tested under two types of difficulty scales and at high and low velocity, with many trials for each scenario.}
\label{tab:stair_table}
\begin{tabular}{|l|l|l|l|l|} \hline 
 Policy&  Terr. Type&  Terr. Difficulty&  Vel. (m/s)&  C.R.\\ \hline 
 MTAC &  Stairs&  50\%&  0.75&  100\%\\ \hline 
 Generalized PPO&  Stairs&  50\%&  0.75&  50\%\\ \hline 
 MTAC&  Stairs&  100\%&  0.75& 
100\%\\ \hline 
 Generalized PPO& Stairs& 100\%& 0.75&30\%\\ \hline 
 MTAC& Stairs& 50\%& 1.75&100\%\\ \hline 
 Generalized PPO& Stairs& 50\%& 1.75&25\%\\ \hline 
 MTAC& Stairs& 100\%& 1.75&82\%\\ \hline 
 Generalized PPO& Stairs& 100\%& 1.75&13\%\\ \hline\end{tabular}
 \vspace{-15pt}
\end{table}
\begin{table}[]
\caption{Bumpy terrain performance metrics. Each model was tested under two types of difficulty scales and at high and low velocity, with many trials for each scenario.}
\label{tab:bumpy_table}
\begin{tabular}{|l|l|l|l|l|} \hline 
 Policy&  Terr. Type&  Terr. Difficulty&  Vel. (m/s)&  C.R.\\ \hline 
 MTAC &  Bumpy&  50\%&  0.75&  100\%\\ \hline 
 Generalized PPO&  Bumpy&  50\%&  0.75&  100\%\\ \hline 
 MTAC&  Bumpy&  100\%&  0.75& 
92\%\\ \hline 
 Generalized PPO& Bumpy& 100\%& 0.75&91\%\\ \hline 
 MTAC& Bumpy& 50\%& 1.75 m/s&75\%\\ \hline 
 Generalized PPO& Bumpy& 50\%& 1.75 m/s&79\%\\ \hline 
 MTAC& Bumpy& 100\%& 1.75 m/s&34\%\\ \hline 
 Generalized PPO& Bumpy& 100\%& 1.75 m/s&13\%\\ \hline\end{tabular}
 \vspace{-15pt}
\end{table}
\begin{table}[]

 \vspace{-15pt}
\caption{Stepped terrain performance metrics. Each model was tested under two types of difficulty scales and at high and low velocity, with many trials for each scenario.}
\label{tab:step_table}
\begin{tabular}{|l|l|l|l|l|} \hline 
 Policy&  Terr. Type&  Terr. Difficulty&  Vel. (m/s)&  C.R.\\ \hline 
 MTAC &  Stepped&  50\%&  0.75&  83\%\\ \hline 
 Generalized PPO&  Stepped&  50\%&  0.75&  41\%\\ \hline 
 MTAC&  Stepped&  100\%&  0.75& 
61\%\\ \hline 
 Generalized PPO& Stepped& 100\%& 0.75&33\%\\ \hline 
 MTAC& Stepped& 50\%& 1.75 m/s&75\%\\ \hline 
 Generalized PPO& Stepped& 50\%& 1.75 m/s&42\%\\ \hline 
 MTAC& Stepped& 100\%& 1.75 m/s&58\%\\ \hline 
 Generalized PPO& Stepped& 100\%& 1.75 m/s&34\%\\ \hline\end{tabular}
 \vspace{-15pt}
\end{table}

%% file: sections/05conc.tex
\section{Conclusion}
We presented a hierarchical reinforcement learning-based controller, MTAC, for rough terrain locomotion in urban search and rescue scenarios for quadruped robots. Without additional sensors or previous knowledge of the simulated terrain, the proposed controller can learn unique and terrain-specialized motions and behaviors to increase the speed and efficiency at which it traverses over different types of terrains and irregularities. The controller is evaluated in multiple test scenes. First, each pre-trained expert is compared against the generalized controller in its native environment type, however, with randomized terrain parameters. Next, the integrated MTAC controller is tested against the previous work in a test map consisting of randomly generated terrains with a goal destination at the end. Each model is evaluated on the time taken to complete its given task, the success rate at which it completes tasks, and the average error between its actual and command base velocity over the time it completes the given task. 
In this methodology, basic domain randomization is used to allow for sim-to-real transfer. However, in this work, we do not focus on transferring the policies learned in simulation to a real robot. In the future, more focus should be applied to improving the existing sim-to-real strategies and applying them to this model. Furthermore, future work should include the use of a gating neural network (GNN) to synthesize the weights of the low-level policies and derive a new policy at each time step. This method would increase the adaptive nature of the generated gaits, and create more natural and complex behaviors. Lastly, future work should explore the use of non-markovian reward scales during training in order to encourage or elicit specific natural behaviors and gaits.